\documentclass[11pt,a4paper]{article}
\usepackage{authblk}
\usepackage[hyperref]{acl2017}
\usepackage{times}
\usepackage{latexsym}
\usepackage[utf8]{inputenc}
\usepackage[T1]{fontenc}
\usepackage[russian,english]{babel}
\usepackage{lmodern}
\usepackage{pgfplots}
\usepackage{url}

\aclfinalcopy 
\def\aclpaperid{194} 

\title{We Built a Fake News \& Click-bait Filter:\\
What Happened Next Will Blow Your Mind!}

\author[1]{Georgi Karadzhov}
\author[1]{Pepa Gencheva}
\author[2]{Preslav Nakov}
\author[1]{Ivan Koychev}

\affil[1]{Sofia University ``St. Kliment Ohridski'', Bulgaria}
\affil[2]{Qatar Computing Research Institute, HBKU, Qatar}
\affil[ ]{\textit{\{georgi.m.karadjov, pepa.k.gencheva\}@gmail.com, }}
\affil[ ]{\textit{pnakov@hbku.edu.qa, koychev@uni-sofia.bg}}

\date{}

\begin{document} 
\maketitle
\begin{abstract}
It is completely amazing! Fake news and click-baits have totally invaded the cyber space. Let us face it: everybody hates them for three simple reasons. Reason \#2 will absolutely amaze you. What these can achieve at the time of election will completely blow your mind! Now, we all agree, this cannot go on, you know, somebody has to stop it. So, we did this research on fake news/click-bait detection and trust us, it is totally great research, it really is! Make no mistake. This is the best research ever!  Seriously, come have a look, we have it all: neural networks, attention mechanism, sentiment lexicons, author profiling, you name it. Lexical features, semantic features, we absolutely have it all. And we have totally tested it, trust us! We have results, and numbers, really big numbers. The best numbers ever! Oh, and analysis, absolutely top notch analysis. Interested? Come read the shocking truth about fake news and click-bait in the Bulgarian cyber space. You won't believe what we have found!
\end{abstract}

\section{Introduction}

Fake news are written and published with the intent to mislead in order to gain financially or politically,
often targeting specific user groups. 
Another type of harmful content on the Internet are the so-called {\it click-baits}, which are distinguished by their sensational, exaggerated, or deliberately false headlines that grab attention and deceive the user into clicking an article with questionable content. 

\noindent While the motives behind these two types of fake news are different, they constitute a growing problem as they constitute a sizable fraction of the online news that users encounter on a daily basis. With the recent boom of Internet, mobile, and social networks, the spread of fake news increases exponentially. Using on-line methods for spreading harmful content makes the task of keeping the Internet clean significantly harder as it is very easy to publish an article and there is no easy way to verify its veracity. Currently, domains that consistently spread misinformation are being banned from various platforms, but this is a rather inefficient way to deal with fake news as websites that specialize in spreading misinformation are reappearing with different domain names. That is why our method is based purely on text analysis,\footnote{An earlier version of the system participated in the \emph{Hack the fake news} hackathon, where it was ranked best in terms of classification accuracy and robustness. See the official results here: \url{https://gitlab.com/datasciencesociety/case\_fake\_news/blob/master/Teams\_Final\_Score.xlsx}} without taking into account the domain name or website's reliability as a source of information. Our work is focused on exploring various stylistic and lexical features in order to detect misleading content, and on experiments with neural network architectures in order to evaluate how deep learning can be used for detecting fake news. Moreover, we created various language-specific resources that could be used in future work on fake news and clickbait detection for Bulgarian, including task-specific word embeddings and various lexicons and dictionaries extracted from the training data.\footnote{The implementation of the final system that we present in this paper is available at \url{https://github.com/lachezarbozhkov/hack\_the\_fake\_news}}


\section{Related Work}
\label{sec:related}

\emph{Trustworthiness and veracity} analytics of on-line statements is an emerging research direction \cite{rowe2009assessing}. 
This includes predicting credibility of information shared in social media \cite{mitra2017parsimonious}, stance classification \cite{zubiaga2016stance} and contradiction detection in rumours \cite{lendvai2016contradiction}.
For example, \newcite{Castillo:2011:ICT:1963405.1963500} 
studied the problem of finding false information about a newsworthy event. They compiled their own dataset, focusing on tweets using a variety of features including user reputation, author writing style, and various time-based features.
\newcite{Canini:2011} analysed the interaction of content and social network structure, 
and \newcite{Morris:2012:TBU:2145204.2145274} studied how Twitter users judge truthfulness.
They found that this is hard to do based on content alone, and instead users are influenced by heuristics such as user name.

\emph{Rumour detection} in social media represents yet another angle of information credibility. \newcite{zubiaga2015analysing} studied how people handle rumours in social media. They found that users with higher reputation are more trusted, and thus can spread rumours among other users without raising suspicions about the credibility of the news or of its source.
\newcite{lukasik-cohn-bontcheva:2015:ACL-IJCNLP} and \newcite{Ma:2015:DRU} used temporal patterns to detect rumours and to predict their frequency,
\newcite{PlosONE:2016} focused on conversational threads,
and \newcite{RANLP2017:factchecking:external} used deep learning to verify claims using the Web as a corpus.

Veracity of information has been also studied in the context of online personal blogs \cite{johnson2007every}, community question answering forums \cite{RANLP2017:credibility:trolls}, 
and political debates \cite{RANLP2017:debates}.


\emph{Astroturfing} and misinformation detection represent another relevant research direction. Their importance is motivated by the strong interest from political science, and research methods are driven by the presence of massive streams of micro-blogging data, e.g., on Twitter \cite{Ratkiewicz:2011:TMS:1963192.1963301}.
While astroturfing has been primarily studied in microblogs such as Twitter, here we focus on on-line news and click-baits instead.

\noindent \emph{Identification of malicious accounts} in social networks is another related research direction. This includes detecting \emph{spam accounts} \cite{almaatouq2016if,mccord2011spam}, \emph{fake accounts} \cite{fire2014friend,cresci2015fame}, \emph{compromised accounts} and \emph{phishing accounts} \cite{adewole2017malicious}. \emph{Fake profile detection} has also been studied in the context of cyber-bullying \cite{galan2014supervised}. A related problem is that of \emph{Web spam detection}, which was addressed as a text classification problem \cite{sebastiani2002machine}, e.g., using spam keyword spotting \cite{dave2003mining}, lexical affinity of arbitrary words to spam content \cite{hu2004mining}, frequency of punctuation and word co-occurrence \cite{li2006combining}. 

\emph{Fake news detection} is most closely related to the present work. While social media have been seen for years as the main vehicle for spreading information of questionable veracity, recently there has been a proliferation of fake news, often spread on social media, but also published in specialized websites.
This has attracted research attention recently. For example, there has been work on studying credibility, trust, and expertise in news communities \cite{mukherjee2015leveraging}.
The credibility of the information published in on-line news portals has been questioned by a number of researchers
\cite{brill2001online,ketterer1998teaching,finberg2002digital}. As timing is crucial when it comes to publishing breaking news, it is simply not possible to double-check the facts and the sources, as is usually standard in respectable printed newspapers and magazines. This is one of the biggest concerns about on-line news media that journalists have \cite{cassidy2007online}.
Finally, \newcite{conroy2015automatic} review various methods for detecting fake news, e.g., using linguistic analysis, discourse, linked data, and social network features.

All the above work was for English. The only work on fact checking for Bulgarian is that of \cite{Hardalov2016}, but they focused on distinguishing serious news from humorous ones. In contrast, here we are interested in finding news that are not designed to sound funny, but to make the reader believe they are real.
Unlike them, we use a deep learning approach.

\section{Fake News \& Click-bait Dataset}
\label{sec:data}
We use a corpus of Bulgarian news over a fixed period of time, whose factuality had been questioned.
The news come from 377 different sources from various domains, including politics, interesting facts and tips\&tricks. 
The dataset was prepared for the \emph{Hack the Fake News} hackathon. It was provided by the Bulgarian Association of PR Agencies\footnote{\url{http://www.bapra.bg/}} and is available in Gitlab\footnote{\url{https://gitlab.com/datasciencesociety/case\_fake\_news/tree/master/data}}. The corpus was automatically collected, and then annotated by students of journalism. 
Each entry in the dataset consists of the following elements:
URL of the original article, date of publication, article heading, article content, a label indicating whether the article is fake or not, and another label indicating whether it is a click-bait.

The training dataset contains 2,815 examples, where 1,940 (i.e., 69\%) are fake news and 1,968 (i.e., 70\%) are click-baits; we further have 761 testing examples. However, there is 98\% correlation between fake news and click-baits, i.e., a model trained on fake news would do well on click-baits and vice versa.
Thus, below we focus on fake news detection only. 

One important aspect about the training dataset is that it contains many repetitions.
This should not be surprising as it attempts to represent a natural distribution of factual vs. fake news on-line over a period of time. As publishers of fake news often have a group of websites that feature the same deceiving content, we should expect some repetition.

In particular, the training dataset contains 434 unique articles with  duplicates. These articles have three reposts each on average, with the most reposted article appearing  45 times. If we take into account the labels of the reposted articles, we can see that if an article is reposted, it is more likely to be fake news. The number of fake news that have a duplicate in the training dataset are 1018 whereas, the number of articles with genuine content that have a duplicate article in the training set is 322. We detect the duplicates based on their titles as far as they are distinctive enough and the content is sometimes slightly modified when reposted.

\noindent \textit{This supports the hypothesis that fake news websites are likely to repost their content.} This is also in line with previous research \cite{Ma:2015:DRU}, which has found it beneficial to find a pattern of how a rumour is reposted over time. 


\section{Method}
\label{sec:method}

We propose a general framework for finding fake news focusing on the text only. We first create some resources, e.g., dictionaries of words strongly correlated with fake news, which are needed for feature extraction. Then, we design features that model a number of interesting aspects about an article, e.g., style, intent, etc. Moreover, we use a deep neural network to learn task-specific representations of the articles, which includes an attention mechanism that can focus on the most discriminative sentences and words.


\subsection{Language Resources}
\label{subsec:language:resources}

As our work is the first attempt at predicting click-baits in Bulgarian, it is organized around building new language-specific resources\footnote{We make these resources freely available in order to promote reproducibility and to enable future research: \url{https://github.com/gkaradzhov/ClickbaitRANLP}} and analyzing the task.

\emph{Word embeddings}: We train 300-dimensional domain-specific word embeddings using word2vec \cite{mikolov2013distributed} on 100,000 Bulgarian news articles from the same sources as the main dataset. The labelled dataset we use in our system is a subset of these articles. Finally, we end up with 207,270 unique words that occur in five or more documents. 
We use these embeddings for text representation, and as an input to our attention-based nevural network. 

\emph{Latent Dirichlet allocation (LDA)}: We use LDA \cite{blei2003latent} in order to build domain-specific topic models, which could be useful for inducing classes of words that signal fake/factual news. The LDA model is trained on the same 100,000 Bulgarian news articles as for training the word embeddings. In our experiments, these LDA classes proved helpful by themselves, but they did not have much to offer on top of the word embeddings. Thus, we ended up not using them in our final system, but we chose to still release them as other researchers might find them useful in the future.

\emph{Fact-checking lexicon}:
Using lexicons of sentiment words has been shown to be very successful for the task of sentiment analysis \cite{Mohammad13}, and we applied the same idea to extract a \emph{fact-checking lexicon}. In particular, we use point-wise mutual information (PMI) to find terms (words, word bi-grams, and named entities) that are highly correlated with the fake/factual news class.
We calculated the PMI scores for uni-grams, bi-grams and on extracted named entities. Table~\ref{PMI-top-words} shows some of the most significant words for the fake news class. We can see in the table some words that grab people attention, but are not very informative by themselves, such as \emph{mysterious} or \emph{phenomenon}. These words are largely context-independent and are likely to remain stable in their usage across different domains and even over an extended period of time. Thus, they should be useful beyond this task and this dataset.

\emph{Other lexicons}: Finally, we create four lexicons that can help to model the difference in language use between fake and factual news articles. In particular, we explored and merged/cleansed a number of on-line resources in order to put together the following lexicons: (\emph{i})~common typos in Bulgarian written text, (\emph{ii})~Bulgarian slang words, (\emph{iii})~commonly used foreign words, and (\emph{iv})~English words with Bulgarian equivalents. We separate the latter two, because of the frequent usage of English words in common language.  
We make these lexicons freely available for future research.


\subsection{Features}
\label{subsec:features}

\subsubsection{Stylometric Features}
Fake news are written with the intent to deceive, 
and their authors often use a different style of writing compared to authors that create genuine content. This could be either deliberately, e.g., if the author wants to adapt the text to a specific target group or wants to provoke some particular emotional reaction in the reader, or unintentionally, e.g., because the authors of fake news have different writing style and personality compared to journalists in mainstream media. Disregarding the actual reason, we use features from author profiling and style detection \cite{rangel2013overview}.

\begin{table}[tbh]
  \begin{center}
    \begin{tabular}{| r | l | r |}
      \hline
      \bf Original word & \bf Translation & \bf PMI \\ \hline
      chemtrails
      & chemtrails 
      & 0.92
      \\ \hline
       \foreignlanguage{russian}{феноменните}
      & the phenomenal  
      & 0.94
      \\ \hline
      \foreignlanguage{russian}{ следете в }
      & follow in 
      & 0.97
       \\ \hline
        \foreignlanguage{russian}{тайнствена}
      & mysterious
      & 0.95
      \\ \hline
       \foreignlanguage{russian}{скрит}
      & hidden
      & 0.84
      \\ \hline
    \end{tabular}
    \caption{Words most strongly associated with the fake news class. \label{PMI-top-words}}
  \end{center}
\end{table}

\emph{Use of specific words that have strong correlation with one of the classes (48 features).} We used the above-described PMI-based fact-checking lexicons to extract features based on the presence of lexicon words in the target article. 
We end up with the following features: 16 for uni-grams + 16 for bi-grams + 16 for named entities, where we have a feature for the sum and also for the average of the word scores for each of the target classes (click-bait, non-click-bait, fake, non-fake), and we had these features separately for the title and for the body of the article.

\emph{Readability index (4 features)}: We calculate standard readability metrics including the type-token ratio, average word length, Flesch–Kincaid readability test \cite{kincaid1975derivation} and Gunning-Fog index \cite{gunning1952technique}. The last two metrics give scores to the text corresponding to the school grade the reader of the target article should have in order to be able to read and understand it easily. These metrics use statistics about the number of syllables, the number of words, and their length. 

\emph{Orthographic features (12 features):} The orthographic features used in our system include: the number of words in the title and in the  content; the number of characters in the title and in the content; the number of specific symbols in the title and in the content, counting the following as symbols \$.!;\#?:-+\@\%\^\&\*(), ; the number of capital letters in the title and in the content; the fraction of capital letters to all letters in the title and in the content; the number of URLs in the content; the overlap between the words from the title and the words of the content, relying on the fact that click-baits tend to have content that does not quite match their title. These features can be very effective for modelling the author's style.

\emph{Use of irregular vocabulary (4 features):} During the initial analysis of our training dataset, we noticed the presence of a high number of foreign words. As it is not common in Bulgarian news articles to use words in another language, we thought that their presence could be a valuable feature to use. One of the reasons for their occurrence might be that they were translated from a foreign resource, or that they were borrowed. We further found that many articles that were labelled as fake news contained a high number of slang words, and we added this as a feature as well. Finally, we have a feature that counts the typos in the text.  

\subsubsection{Lexical Features}

General lexical features are often used in natural language processing as they are somewhat task-independent and reasonably effective in terms of classification accuracy. In our experiments, we used TF.IDF-based features over the title and over the content of the article we wanted to classify. We had these features twice -- once for the title and once for the the content of the article, as we wanted to have two different representations of the same article. Thus, we used a total of 1,100 TF.IDF-weighted features (800 content + 300 title), limiting the vocabulary to the top 800 and 300 words, respectively (which occurred in more than five articles). We should note that TF.IDF features should be used with caution as they may not remain relevant over time or in different contexts without retraining. 

\subsubsection{Grammatical Features}

The last type of hand-crafted features that we used are the grammatical features. First, we evaluate how often stop words are used in the content of the article. Extensive usage of stop words may indicate irregularities in the text, which would be missed by the above features. Additionally, we extract ten coarse-grained part-of-speech tags from the content of the article and we use part-of-speech occurrence ratios as features. This makes a total of twenty features, as we have separate features for the title and for the contents.


\subsubsection{Semantic Features}

All the above features are hand-crafted, evaluating a specific text metric or checking whether specific words highly correlate with one of the classes. However, we lack features that target the semantic representation of the text itself. Thus, we further use two types of word representations.


\emph{Word embeddings (601 features).} As we said above, we trained domain-specific word embeddings.
In order to incorporate them as features, we calculate the average vector for the title and separately for the content of the news article. We end up with two 300-dimensional embedding representations of the semantics of the articles, which we use as 300+300=600 features. We also compute the cosine similarity between the average vector of the title and the average vector of the content, because we believe that this is a highly indicative measure for at least click-bait articles, whose content differs from what their title says.

\emph{Task-specific embeddings.} As a more advanced representation, we feed the text into an attention-based deep neural network, which we train to produce a task-specific embedding of the news articles. The network is designed to recognize words and sentences that contribute to the click-bait class attribution. The architecture is described in details in Section \ref{subsec:attention}

\subsection{Some Features that we Ignored}

As we mentioned above, our method is purely text-based. Thus, we ignored the publishing date of the article. In future work, it could be explored as a useful piece of information about the credibility of the article, as there is interesting research in this direction \cite{Ma:2015:DRU}. We also disregarded the article source (the URL) because websites that specialize in producing and distributing fake content are often banned and then later reappear under another name. We recognize that the credibility of a specific website could be a very informative feature, but, for the sake of creating a robust method for fake news detection, our system relies only on the text when predicting whether the target article is likely to be fake. We describe our features in more detail below.

\subsection{Model}
\label{subsec:model}

Our framework for fake news detection is comprised of two components, which are used one after the other. First, we have an attention-based deep neural network model, which focuses on the segments of the text that are most indicative of the target class identification, and as a side effect learns task-specific representations of the news articles. We extract these representations from the last hidden layer in the network, and we feed it to the SVM classifier together with the hand-crafted features.

\subsubsection{Attention Mechanism}
\label{subsec:attention}

The attention network \cite{DBLP:journals/corr/HermannKGEKSB15}, \cite{yang2016hierarchical} is a powerful mechanism, inspired by the human ability to spot important sections in images or text. We adopt the approach used in \cite{rocktaschel2015reasoning} and employ an attention neural networks to build attention over the text of a piece of news with respect to the title it has. As far as it is in the nature of click-baits to have titles that are different from the text of the news, the attentional layers of the neural network should spot when the two texts talk about the same thing and when they are not corresponding or accurate. We implemented the attention mechanism using Keras \cite{chollet2015keras} with the Tensorflow back-end \cite{tensorflow2015-whitepaper}. 

\begin{figure}[h!]
\center
\caption{The architecture of our hierarchical attention deep neural network for click-bait news detection.}
\includegraphics[width=7cm, height=11cm]{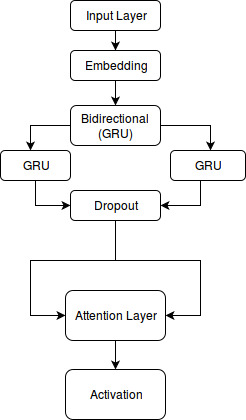}
\label{pic:attention}
\end{figure}

The architecture of the network with attention layers is shown in Figure~\ref{pic:attention}. Our neural model is based on Gated Recurrent Units (GRUs). GRUs are gating mechanism in RNNs which provide the ability to learn long-term dependencies and were first introduced in \cite{cho2014properties}.  Given the document embedding, the GRUs build representations using input and forget gates, which help storing the valuable information through time. They build embeddings of the title and the text of the news, where at each step the unit has information only about the output from the previous step. This can be considered as a drawback, as far as we would considerably benefit if each step could construct its decision based not only on the previous step's output, but on all of the words that were processed so far. To improve this, the attention layer, for each step in the text sequence, uses the output of the steps in the title sequence. Thus, the layer learns weights, designating the strength of the relatedness between each word in the title and each word in the content. 

For the neural network, we take the first 50 symbols of the title and the content of the news, which we choose after experiments. We train the neural network for 20 epochs and the final classification is derived with sigmoid activation. The optimizer used for the training is Adam optimizer. We feed the neural network with the embedding of the words we built earlier with word2vec.

As we will see below, the neural network is inferior in terms of performance to a feature-rich SVM (even though it performs well above the baseline). This is because it only has access to word embeddings, and does not use the manually-crafted features. Yet, its hidden layer represents a 128-dimensional task-specific embedding of the input article,
and it turns out that using it as a list of 128 features in the SVM classifier yields even further great improvement, as we will see below. In this way, we combine a deep neural network with an attention mechanism with kernel-based SVM.


\begin{table}[tbh]
\begin{center}
  \begin{tabular}{@{}|l|cccc|@{}}
    \hline
    \bf Features & \bf P & \bf R & \bf F1 & \bf Acc \\ \hline
    Lexical & 75.53 & 74.59 & 75.02 & 79.89\\ \hline
    Stylometric &74.35&65.99&67.68&77.52\\\hline
    Grammatical & 73.23 & 50.60 & 42.99 & 71.48\\\hline
    Embeddings &61.48&53.95&51.67& 71.22\\\hline
  \end{tabular}
\end{center}
\caption{Performance of the individual groups of hand-crafted features.\label{individual-features}}
\end{table}

\subsubsection{SVM}

We feed the above-described hand-crafted features together with the task-specific embeddings learned by the deep neural neural network (a total of 1,892 attributes combined) into a Support Vector Machines (SVM) classifier \cite{Cortes1995}. SVMs have proven to perform well in different classification settings, including in the case of small and noisy datasets. 

\section{Experiments and Evaluation}
\label{sec:experiments}

We trained on the 2,815 training examples, and we tested on the 761 testing ones. The test dataset was provided apart from the training one, thus we didn't have to partition the original dataset to receive a testing one. The validation of the models was performed on a randomly chosen subset of sentences - one fifth of the original set.  
We scaled each feature individually by its maximum absolute value to end up with each feature having values in the [0;1] interval.
We used an RBF kernel for the SVM, and we tuned the values of $C$ and $\gamma$ using cross-validation.
We trained the neural network using RMSProp \cite{tieleman2012lecture} with a learning rate of 0.001 and mini-batches of size 32, chosen by performing experiments with cross-validation .
We evaluated the model after each epoch and we kept the one that performed best on the development dataset. 

Table \ref{individual-features} shows the performance of the features in groups as described in Section~\ref{subsec:features}. We can see that, among the hand-crafted features, the lexical features yield the best results, i.e., words are the most indicative features. The good results of the stylometric features indicate that the intricacies of language use are highly discriminative. The next group is the one with the grammatical features, which shows good performance in terms of Precision. The last one are the embedding features, which although having low individual performance, contribute to the overall performance of the system as shown in next paragraph.

Evaluating the final model, we set as a baseline the prediction of the majority class, i.e., the fake news class. This baseline has an F1 of 41.59\% and accuracy of 71.22\%. The performance of the built models can be seen in Table~\ref{models}. Another stable baseline, apart from just taking the majority class, is the TF.IDF bag-of-words approach, which sets a high bar for the general model score. We then observe how much the attention mechanism embeddings improve the score (AttNN). Finally, we add the hand-crafted features (Feats), which further improve the performance. From the results, we can conclude that both the attention-based task-specific embeddings and the manual features are important for the task of finding fake news.

\begin{table*}[tbh]
\begin{center}
  \begin{tabular}{|l|cccc|}
    \hline
    \bf Feature Group & \bf P & \bf R & \bf F1 & \bf Acc \\ \hline
    Baseline & 35.61 & 50.00 & 41.59& 71.22\\ \hline
    TF.IDF & 75.53 & 74.59 & 75.02 & 79.89 \\ \hline
    AttNN & 78.52 & 78.74 & 78.63 & 81.99\\ \hline
    TF.IDF \&AttNN & 79,89 & 79.40 & 79.63 & 83.44 \\ \hline
    TF.IDF \&Feats \&AttNN & 80.07 & 79.49 & 79.77 & 83.57 \\
    \hline
  \end{tabular}
\end{center}
\caption{Performance of different models.\label{models}}
\end{table*}



\section{Conclusion and Future Work}
\label{sec:future}

We have presented the first attempt to solve the fake news problem for Bulgarian. 
Our method is purely text-based, and ignores the publication date and the source of the article.
It combines task-specific embeddings, produced by a two-level attention-based deep neural network model, with manually crafted features (stylometric, lexical, grammatical, and semantic), into a kernel-based SVM classifier.
We further produced and shared a number of relevant language resources for Bulgarian, which we created for solving the task.



The evaluation results are encouraging and suggest the potential applicability of our approach in a real-world scenario. 
They further show the potential of combining attention-based task-specific embeddings with manually crafted features. 
An important advantage of the attention-based neural networks is that the produced representations can be easily visualized and potentially interpreted as shown in \cite{DBLP:journals/corr/HermannKGEKSB15}. We consider the implementation of such visualization as an important future work on the task.

\section*{Acknowledgements} 
We would like to thank Lachezar Bozhkov, who was part of our team in the \emph{Hack the Fake News} hackathon, for his insight. This work is supported by the NSF of Bulgaria under Grant No. DN-02/11/2016 - ITDGate.

\bibliographystyle{acl_natbib}
\bibliography{acl2017}

\end{document}